\documentclass{llncs}
\usepackage{capt-of}
\usepackage{booktabs,caption,fixltx2e}
\usepackage[flushleft]{threeparttable}
\usepackage{graphicx}
\usepackage{makeidx} 
\usepackage{amsmath}
\usepackage{multirow}

\usepackage{amsfonts}%
\usepackage{amssymb}%
\usepackage[english,oldcommands,plain,lined]{algorithm2e}
\usepackage{graphics}
\usepackage{subfig}
\usepackage[labelfont=bf]{caption}
\usepackage[font={small}]{caption}
\usepackage{transparent}
\usepackage{color}
\usepackage{breqn}
\usepackage{epstopdf}
\usepackage{wrapfig}

\usepackage{color, colortbl}
\definecolor{LightCyan}{rgb}{0.88,1,1}
 \usepackage{cite}
\usepackage[font=small,labelfont=bf]{caption}
\usepackage{enumitem}
\usepackage{multirow}
\usepackage{hyperref}
\newcolumntype{P}[1]{>{\centering\arraybackslash}p{#1}}

\begin{document}

\title{Inherent Brain Segmentation Quality Control from Fully ConvNet Monte Carlo Sampling}

\author{Abhijit Guha Roy\inst{1,2}, Sailesh Conjeti\inst{3}, Nassir Navab\inst {2,4}, Christian Wachinger\inst{1}}

\vspace{-2mm}
\institute{
$^1$Artificial Intelligence in Medical Imaging (AI-Med), KJP, LMU M\"{u}nchen, Germany.\\
$^2$Computer Aided Medical Procedures, Technische Universit\"{a}t M\"{u}nchen, Germany.\\
$^3$German Center for Neurodegenerative Diseases (DZNE), Bonn, Germany.\\
$^4$Computer Aided Medical Procedures, Johns Hopkins University, USA.}

\maketitle 
\vspace{-2mm}
\begin{abstract}
We introduce inherent measures for effective quality control of brain segmentation based on a Bayesian fully convolutional neural network, using model uncertainty. Monte Carlo samples from the posterior distribution are efficiently generated using dropout at test time. Based on these samples, we introduce next to a voxel-wise uncertainty map also three metrics for structure-wise uncertainty. We then incorporate these structure-wise uncertainty in group analyses as a measure of confidence in the observation. Our results show that the metrics are highly correlated to segmentation accuracy and therefore present an inherent measure of segmentation quality. Furthermore, group analysis with uncertainty results in effect sizes closer to that of manual annotations. The introduced uncertainty metrics can not only be very useful in translation to clinical practice but also provide automated quality control and group analyses in processing large data repositories.
\end{abstract}

\section{Introduction}
\label{sec:intro}
Magnetic resonance imaging (MRI) delivers high-quality, \emph{in-vivo} information about the brain. Whole-brain segmentation~\cite{freesurfer2002,fsl2012} provides imaging biomarkers of neuroanatomy, which form the basis for tracking structural brain changes associated with aging and disease. Despite efforts to deliver robust segmentation results across scans from different age groups, diseases, field strengths, and manufacturers, inaccuracies in the segmentation outcome are inevitable~\cite{Mindcontrol}. A manual quality assessment is therefore recommended before continuing with the analysis. However, the manual assessment is not only time consuming, but also subject to inter- and intra-rater variability. 

The underlying problem is that most segmentation algorithms provide results without a measure of confidence or quality. Bayesian approaches are an alternative, because they do not only provide the mode (i.e., the most likely segmentation) but also the posterior distribution. However, most Bayesian approaches use point estimates in the inference, whereas marginalization over parameters has only been proposed in combination with Markov Chain Monte Carlo sampling~\cite{iglesias2013improved}  or the Laplace approximation~\cite{wachinger2015seg}. While sampling-based approaches incorporate fewer assumptions, they are computationally intense and have so far only been used for the segmentation of substructures but not the whole-brain~\cite{iglesias2013improved}. 

Recent advances in Bayesian deep learning enabled approximating the posterior distribution by dropping out neurons at test time~\cite{gal2016}. This does not require any additional parameters and is achieved by sampling from the Bernoulli distribution across the network weights. In addition, this approach enables to represent uncertainty in deep learning without sacrificing accuracy or computational complexity, allowing for fast Monte Carlo sampling. This concept of uncertainty was later extended for semantic segmentation within fully convolutional neural networks (F-CNN)~\cite{baySegNet2017} providing a pixel-wise uncertainty estimation. At the same time, F-CNNs started to achieve state-of-the-art performance for whole-brain segmentation, while requiring only seconds for a 3D volume~\cite{ecb2017,quickNat2018}. 

In this work, we propose inherent measures of segmentation quality based on a Bayesian F-CNN for whole-brain segmentation. To this end, we extend the F-CNN architecture~\cite{quickNat2018} with dropout layers, which allows for highly efficient Monte Carlo sampling. From the samples, we compute the the voxel-wise segmentation uncertainty and introduce three metrics for quantifying uncertainty per brain structure. We show that these metrics are highly correlated with the segmentation accuracy and can therefore be used to predict segmentation accuracy in absence of ground truth. Finally, we propose to effectively use the uncertainty estimates as quality control measures in large-scale group analysis to estimate reliable effect sizes. We believe that uncertainty measures are not only essential for the translation of quantitative measures to clinical practice but also provide automated quality control and group analyses in large data repositories.

\noindent
\textbf{Prior Art: }
Evaluating segmentation performance without ground truth has been studied in medical imaging before. In early work, the common agreement strategy (STAPLE) was used to evaluate classifier performance for segmenting brain scans into WM, GM and CSF~\cite{bouix2007}. In another approach, features corresponding to a segmentation map were used to learn a separate regressor for predicting the Dice score~\cite{kohlberger2012}. Recently, the reverse classification accuracy was proposed, which involves training a separate classifier on the segmentation outcome of the method to evaluate, serving as pseudo ground truth~\cite{valindria2017}. In contrast to these previous approaches, we provide a quality measure that is \emph{inherently} computed within the segmentation framework, derived from model uncertainty and does therefore not require training a second, independent classifier for evaluation, which itself may be subject to prediction errors.

\begin{figure}[t]
\centering
\includegraphics[width=0.95\textwidth]{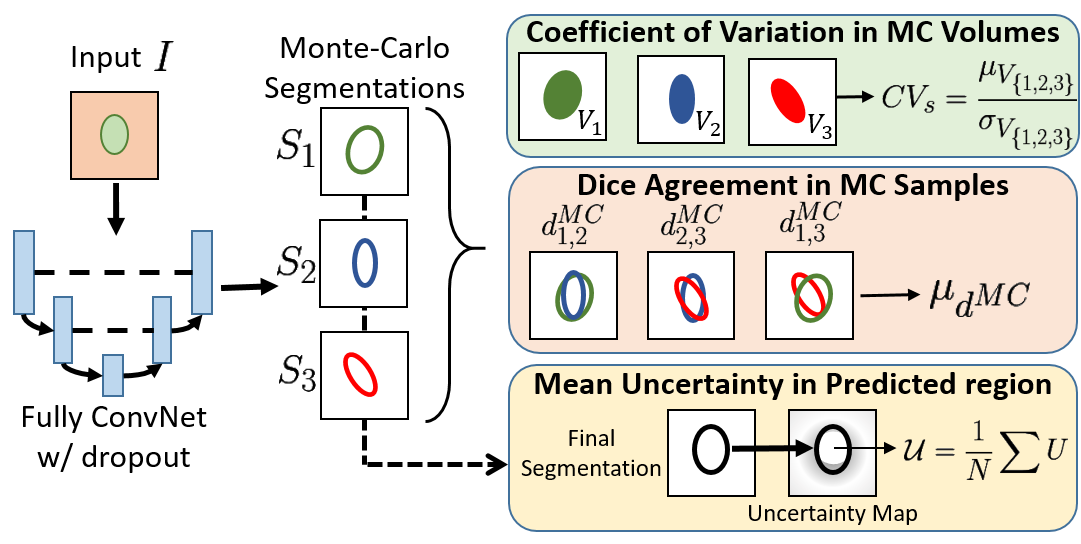}
\vspace{-2mm}
\caption{A single input scan results in different Monte Carlo (MC) segmentations ($S_1,S_2,S_3$) based on different dropouts in the fully ConvNet.
The samples are used to estimate  three variants of structure-wise uncertainty. 
The final segmentation is the average of the MC samples, used in the third variant.}
\label{fig:graphAbs}
\end{figure}

\section{Method}
\label{sec:methodology}

\noindent
\textbf{Bayesian Inference: }
We employ dropout~\cite{dropout} to create a probabilistic encoder-decoder network, which approximates probabilistic neuron connectivity similar to a Bayesian neural network (BNN)~\cite{gal2016}. Dropout is commonly used in training and then turned-off at testing time. 
By using dropout also at testing, we can sample from the posterior distribution of the model. We modify the architecture in~\cite{quickNat2018} by \emph{inserting dropout layers after every encoder and decoder block} with a dropout rate of $q$. 

A given input $I$ is feed-forwarded $N$ times with different dropped out neurons, generating $N$ different Monte Carlo (MC) samples of segmentation $\{S_1, \cdots S_N\}$. This inference strategy is similar to variational inference in BNNs, assuming a Bernoulli distribution over the weights~\cite{gal2016}. The final probability map is given by computing the average over MC probability maps. We set the dropout rate to $q=0.2$ and produce $N=15$ MC samples ($<$ 2 minutes), after which performance saturates.
We pre-train the network on $581$ volumes of the IXI dataset\footnote{http://brain-development.org/ixi-dataset/} with FreeSurfer~\cite{freesurfer2002} segmentations and subsequently fine-tune on $15$ of the 30 manually annotated volumes from the Multi-Atlas Labelling Challenge (MALC) dataset~\cite{OasisMultiAtlas}. This trained model is used for all our experiments. In this work, we segment $33$ cortical and sub-cortical structures.

\subsection{Uncertainty Measures}
\noindent
\textbf{1. Voxel-wise Uncertainty: }
The model uncertainty $U_s$ for a given voxel $\mathbf{x}$, for a specific structure $s$ is estimated as entropy over all $N$ MC probability maps~$p_s$

\begin{equation}
U_s(\mathbf{x}) = - \sum_{i=1}^N p_s^i(\mathbf{x}) \log(p_s^i(\mathbf{x})).
\label{eq:unc}
\end{equation}

\noindent 
The voxel-wise uncertainty is the sum over all structures, $U = \sum_s U_s$. Voxels where uncertainty is low (i.e. entropy is low) receive the same predictions, in spite of different neurons being dropped out. 

\noindent
\textbf{2. Structure-wise Uncertainty: }
For many applications, it is helpful to have an uncertainty measure per brain structure. 
We propose three different strategies for computing structure-wise uncertainty from MC segmentations, illustrated in  Fig.~\ref{fig:graphAbs} for  $N=3$ MC samples. 

\noindent
\textbf{Type-1: }
We measure the variation of the volume across the MC samples. We compute the coefficient of variation $CV_s = \frac{\sigma_s}{\mu_s}$ for a structure $s$, with mean~$\mu_s$ and standard deviation~$\sigma_s$ of MC volume estimates. 
Note that this estimate is agnostic to the size of the structure.

\noindent
\textbf{Type-2: }We use the overlap between samples as a measure of uncertainty. 
To this end, we compute the average Dice score over all  pairs of MC samples 
\begin{equation}
d_s^{MC} = E \left[ \{ Dice((S_i==s), (S_j==s)) \}_{i \neq j} \right].
\end{equation}
\noindent

\noindent
\textbf{Type-3: }We define the uncertainty for a structure $s$ as mean voxel-wise uncertainty over the voxels which were labeled as $s$,  $\mathcal{U}_s = E \left[  \{ U(\mathbf{x}) \}_{\mathbf{x}\in \{ S==s \}}  \right]$.

Note that $d_s^{MC}$ is directly related to segmentation accuracy, while $\mathcal{U}_s$ and $CV_s$ are inversely related to accuracy.

\subsection{Segmentation Uncertainty in Group Analysis }
We propose to integrate the structure-wise uncertainty in group analysis. 
To this end, we solve a weighted linear regression model with weight $w_i$ for subject $i$
\begin{equation}
\hat{\boldsymbol{\beta}} = \arg \min \sum_i \omega_i (V_i - \mathbf{X}_i \boldsymbol{\beta}^\top)^2
\label{eq:weightedAnalysis}
\end{equation}
with design matrix $\mathbf{X}$, vector of coefficients $\boldsymbol{\beta}$, and brain structure volume  $V_i$. 
We use the first two types of structure-wise uncertainty and set the weight $\omega_i$ to $\frac{1}{CV_s}$ or $\frac{1}{1-d_s^{MC}}$. 
Including  weights in  linear regression increases its robustness as scans with reliable segmentation are emphasized.  
Setting all weights to a constant results in standard regression. 
In our experiments, we set 
\begin{equation}
\mathbf{X}_i = [1, A_i, S_i, D_i]  \quad \quad    \boldsymbol{\beta} = [\beta_0, \beta_A, \beta_{S}, \beta_{D}]
\end{equation}
with age $A_i$, sex $S_i$ and diagnosis $D_i$ for subject $i$. 
Of particular interest is the regression coefficient $\beta_{D}$, which estimates the effect of diagnosis on the volume of a brain structure $V$.

\section{Experimental Results}
\label{sec:exp}
\noindent
\textbf{Datasets: } We test on the 15 volumes of the \textbf{MALC} dataset~\cite{OasisMultiAtlas} that were not used for training. Further, we deployed the model on un-seen scans across 3 different datasets not used for training : 
(i) \textbf{ADNI-29}: The dataset consists of 29 scans from ADNI dataset~\cite{adni2008}, with a balanced distribution of Alzheimer's Disease (AD) and control subjects, and scans acquired with 1.5T and 3T scanners. The objective is to observe uncertainty changes due to variability in scanner and pathologies. 
(ii) \textbf{CANDI-13}: The dataset consists of 13 brain scans of children (age 5-15) with psychiatric disorders, part of the CANDI dataset~\cite{candi2012}. The objective is to observe changes in uncertainty for data with age range not included in training. 
(iii) \textbf{IBSR-18}: The dataset consist of 18 scans publicly available at \url{https://www.nitrc.org/projects/ibsr}. The objective is to see the sensitivity of uncertainty with low resolution and poor contrast scans.
Note that the training set (MALC) did not contain scans with AD or scans from children. 
Manual segmentations for MALC, ADNI-29, and CANDI-13 were provided by Neuromorphometrics, Inc.\footnote{http://Neuromorphometrics.com/}

\begin{table}[t]
\scriptsize
\centering
\caption{Results on 4 different datasets with global Dice scores and correlation of Dice scores with 3 types of uncertainty.}
  \begin{tabular}{|p{0.64in}|P{0.90in}| P{0.65in}|P{0.40in}|P{0.40in}|P{0.40in}|}
    \hline
     Datasets & Mean Dice Score & Mean & \multicolumn{3}{c|}{Corr($\cdot$, DS)}  \\
      & (DS) & $CV_s$ & $\mathcal{U}_s$ & $CV_s$ & $d_s^{MC}$ \\
    \hline
    \textbf{MALC-15} & $\mathbf{0.88}\pm0.02$ & $0.38$ & $-0.85$ & $-0.81$ & $\mathbf{0.86}$  \\ 
    \textbf{ADNI-29} & $0.83\pm0.02$ & $0.46$ & $-0.72$ & $-0.71$ & $\textbf{0.78}$  \\ 
    \textbf{CANDI-13} & $0.81\pm0.03$ & $0.54$ & $-0.84$ & $-0.86$ & $\mathbf{0.90}$ \\ 
    \textbf{IBSR-18} & $0.81\pm0.02$ & $0.57$ & $-0.76$ & $-0.76$ & $\mathbf{0.80}$ \\ \hline
  \end{tabular}
  \label{tab:res}
\end{table}

\begin{figure}[t]
\centering
\includegraphics[width=0.9\textwidth]{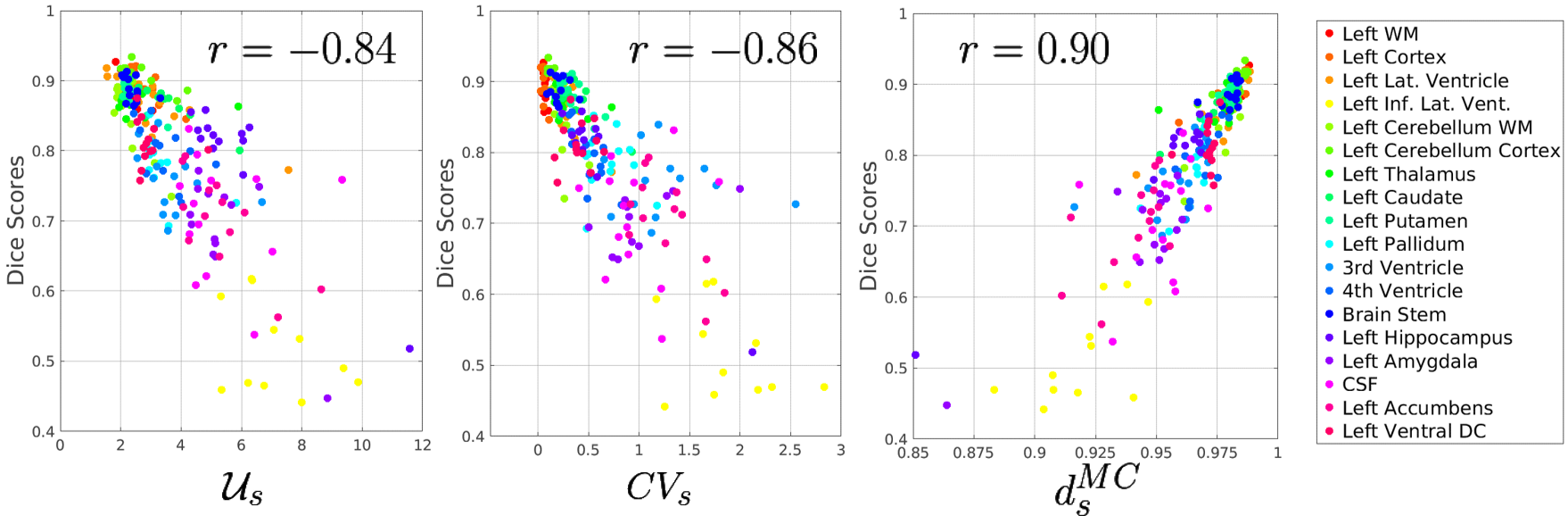}
\vspace{-2mm}
\caption{Scatter plot of three types of uncertainty and Dice scores  on CANDI-13 dataset (one dot per scan and structure), with their corresponding correlation coefficient ($r$). For clarity, structures only on the left hemisphere are shown. }
\label{fig:scatterplot}
\end{figure}

\noindent
\textbf{Quantitative Analysis: } 
To quantify the performance of the uncertainty in predicting the segmentation accuracy, we compute the correlation coefficient between the Dice scores and the three types of structure-wise uncertainty. 
Fig.~\ref{tab:res} reports the correlations for all 4 test datasets, together with the Dice score of the inferred segmentation.  
Firstly, we observe that the segmentation accuracy is highest on MALC and that the accuracy drops ($5-7\%$) for other datasets (ADNI, CANDI, IBSR). 
This decrease in performance is to be expected when transferring the model to other datasets and is also reflected in the uncertainty estimate (Mean $CV_s$). 
Secondly, for the three measures of structure-wise uncertainty, the Dice agreement in MC samples $d_s^{MC}$ shows highest correlations across all datasets. 
The overall high correlation for $d_s^{MC}$ indicates that it is a suitable proxy for measuring segmentation accuracy without the presence of ground truth annotations. 
Fig.~\ref{fig:scatterplot} shows scatter plots for the three uncertainty variants with respect to actual Dice score on CANDI-13. 

\begin{figure}[t]
\centering
\includegraphics[width=0.85\textwidth]{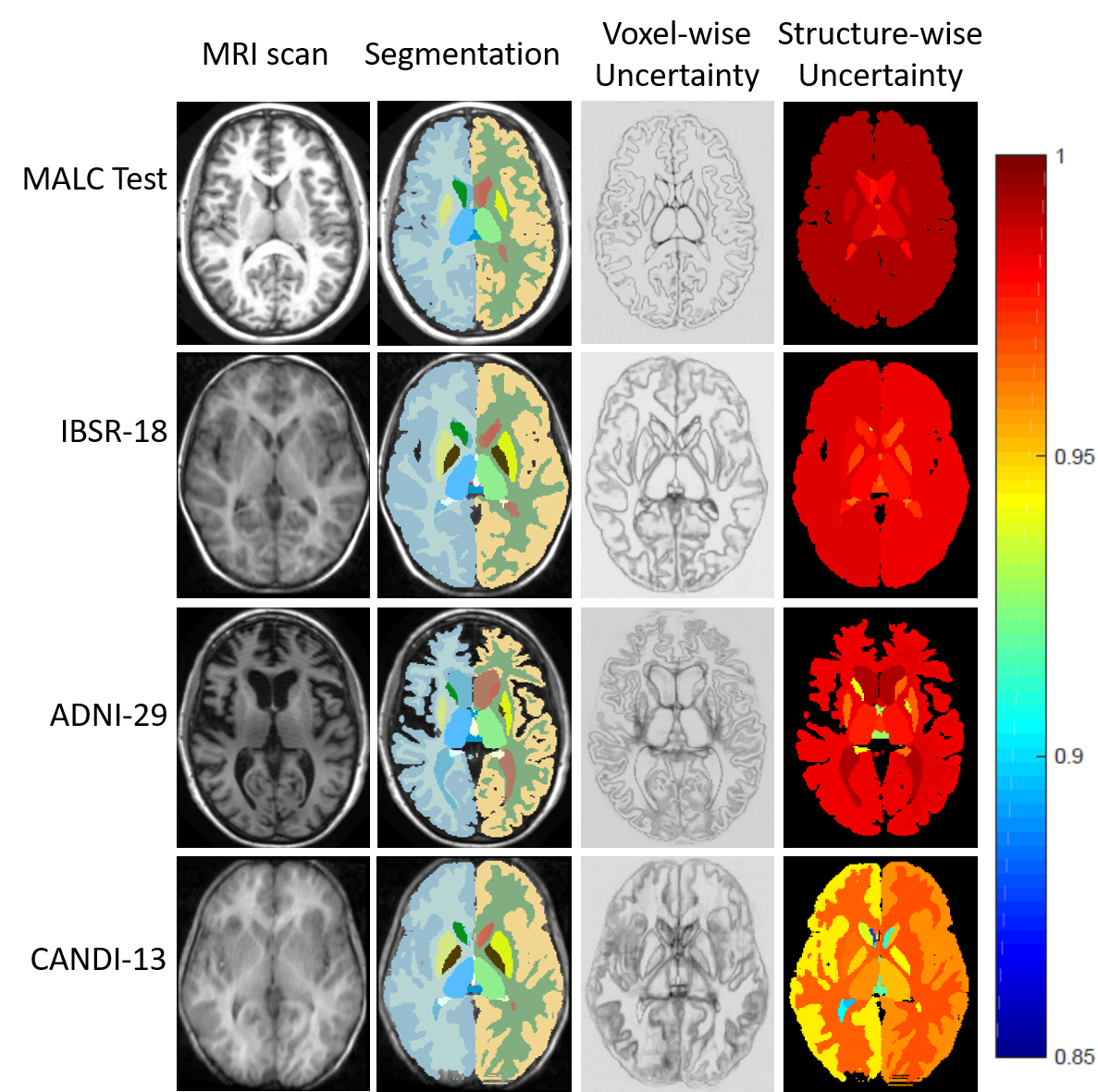}
\caption{Results of 4 different cases, one from each dataset, corresponding to the worst Dice score. The MRI scan, segmentation, voxel-wise uncertainty and structure-wise uncertainty ($d_s^{MC}$) are presented. Red in the heat map indicates high reliability in segmentation, while blue indicates poor segmentation.}
\label{fig:results}
\end{figure}

\noindent
\textbf{Qualitative Analysis: }
Fig.~\ref{fig:results} illustrates qualitative results with  MRI scan, segmentation, voxel-wise uncertainty map and structure-wise uncertainty ($d_{MC}$) heat map. 
In the heat map, red indicates higher reliability in segmentation. 
The first row shows results on a test sample from the MALC dataset, where segmentation is good with high reliability in prediction. 
The second row presents the scan with worst performance on IBSR-18 dataset, consisting of poor contrast with prominent ringing artifacts. 
Its voxel-wise and structure-wise uncertainty maps shows less reliability in comparison to MALC. 
The third row presents the scan with worst performance in ADNI-29, a subject of age 95 with severe AD. Prominent atrophy in cortex along with enlarged ventricles are visible in the MRI scan, with ringing artifacts at the top. Its $d_s^{MC}$ heat maps shows higher uncertainty in some subcortical structures with brighter shades. The last row presents the MRI scan with the worst performance on CANDI-13 dataset, a subject of age 5 with high motion artifact together with poor contrast. Its voxel-wise uncertainty is higher in comparison to others, with dark patches prominent in subcortical regions. 
The heat map shows the lowest confidence for this scan, in comparison to other results.

\noindent
\textbf{Uncertainty for Group Analysis: }
In this section, we evaluate the integration of structure-wise uncertainty in group analyses. 
First, we perform  group analysis on ADNI-29 with 15 control and 14 AD subjects. 
We focus our analysis on most prominent AD biomarkers, the volume of hippocampus and lateral ventricles~\cite{Thompson2004}.
Table~\ref{tab:groupAnalysis} reports the regression coefficient and p-value for diagnosis ($\beta_{D}$, $p_{D}$). 
The coefficient is computed by solving Eq.~\ref{eq:weightedAnalysis}, where we use two types of uncertainty ($CV_s$, $d_s^{MC}$) and compare to  normal regression. 
Although the dataset is small, it comes with ground truth annotations and therefore allows for estimating the actual $\beta_{D}$. 
Comparing, we observe that both versions of weighted regression results in $\beta_{D}$  closer to the actual effect in comparison to normal regression. Also, we note that $CV_s$ provides a better weighting than $(1-d_s^{MC})$. 
Next, we perform group analysis on the ABIDE-I dataset~\cite{abide2014} consisting of $1,112$ scans, with $573$ normal subjects and $539$ subjects with autism. The dataset is collected from 20 different sites with a high variability in scan quality. 
To factor out changes due to site, we added site as a covariate in Eq.~\ref{eq:weightedAnalysis}. 
We report $\beta_{D}$ with corresponding p-values for the volume of brain structures that have recently been associated to autism in a large ENIGMA study~\cite{autism}. We compare uncertainty weighted regression to normal regression, and include  robust regression with Huber norm. 
$CV_s$ provides the highest effect sizes, followed by $(1-d_s^{MC})$. 
Strikingly, uncertainty weighted regression results in significant associations to autism, identical to~\cite{autism}, whereas normal regression is only significant for amygdala.

\begin{table}[t]
\scriptsize
\centering
\caption{Results of group analyses on ADNI-29 and ABIDE datasets with pathologies (Alzheimer's and autism), with and without using uncertainty.}
  \begin{tabular}{|p{0.85in}|P{0.40in}|P{0.40in}|P{0.40in}|P{0.40in}|P{0.40in}|P{0.40in}|P{0.40in}|P{0.40in}|}
      \hline
    & \multicolumn{8}{c|}{\textbf{ADNI-29}}  \\ 
     \textbf{AD Biomarkers} & \multicolumn{2}{c|}{Ground Truth} & \multicolumn{2}{c|}{Normal Regression} & \multicolumn{2}{c|}{$CV_s$} &
      \multicolumn{2}{c|}{$d_s^{MC}$} \\ 

      & $\beta_{D}$ & $p_{D}$ & $\beta_{D}$ & $p_{D}$  & $\beta_{D}$ & $p_{D}$  & $\beta_{D}$ & $p_{D}$ \\
    \hline
    \textbf{Hippocampus} & $1.16$ & $0.0010$ & $1.26$ & $0.0002$ & $1.21$ & $0.0002$ & $1.25$ & $0.0002$ \\ 
    \textbf{Lat. Ventricle} & $-0.15$ & $0.6658$ & $-0.19$ & $0.5826$ & $-0.15$ & $0.6650$ & $-0.16$ & $0.6342$ \\ 
    \hline
    & \multicolumn{8}{c|}{\textbf{ABIDE}}  \\ 
     \textbf{Autism} & \multicolumn{2}{c|}{Normal Regression} & \multicolumn{2}{c|}{Robust Regression} & \multicolumn{2}{c|}{$CV_s$} &
      \multicolumn{2}{c|}{$d_s^{MC}$} \\ 

      \textbf{Biomarkers}& $\beta_{D}$ & $p_{D}$ & $\beta_{D}$ & $p_{D}$  & $\beta_{D}$ & $p_{D}$  & $\beta_{D}$ & $p_{D}$ \\
    \hline
    \textbf{Amygdala} & $-0.14$ & $0.0140$ & $-0.07$ & $0.0499$ & $-0.32$ & $0.0001$ & $-0.27$ & $0.0001$ \\ 
    \textbf{Lat. Ventricles} & $-0.01$ & $0.8110$ & $-0.05$ & $0.1294$ & $-0.38$ & $0.0089$ & $-0.19$ & $0.0843$ \\ 
    \textbf{Pallidum} & $-0.07$ & $0.2480$ & $-0.01$ & $0.8727$ & $-0.40$ & $0.0051$ & $-0.28$ & $0.0165$ \\
    \textbf{Putamen} & $-0.07$ & $0.2186$ & $-0.01$ & $0.8125$ & $-0.43$ & $0.0035$ & $-0.39$ & $0.0057$ \\
    \textbf{Accumbens} & $-0.08$ & $0.1494$ & $-0.03$ & $0.4386$ & $-0.21$ & $0.0013$ & $-0.17$ & $0.0031$ \\ \hline
  \end{tabular}
  \label{tab:groupAnalysis}
\end{table}

\vspace{-2mm}
\section{Conclusion}
\label{sec:conc}
\vspace{-2mm}
We introduced a Bayesian F-CNN model for whole-brain segmentation that produces MC samples by using dropout at test time. Based on the samples, we introduced metrics for quantifying structure-wise uncertainty. We show a high correlation with segmentation accuracy of these metrics on 4 out-of-sample datasets, thus providing segmentation quality. In addition, we proposed to integrate the confidence in the observation into group analysis, yielding improved effect sizes.


\noindent
\textbf{Acknowledgement:} We thank SAP SE and the Bavarian State Ministry of Education, Science and the Arts in the framework of the Centre Digitisation.Bavaria (ZD.B) for funding and the NVIDIA corporation for GPU donation.

\end{document}